\documentclass[sigconf]{acmart}

\usepackage{mathrsfs}
\usepackage{multirow}
\usepackage{slashbox}
\usepackage{booktabs}
\usepackage{CJK}
\usepackage{color}
\usepackage{psfrag}    
\usepackage{subfigure} 
\usepackage{url}       
\usepackage{stfloats}  
\usepackage{amsmath}   
\usepackage{array}
\usepackage{breqn}
\usepackage{amsfonts}
\usepackage{helvet}
\usepackage{courier}
\usepackage{amssymb}
\usepackage{{inputenc}}
\usepackage{balance}
\usepackage{multirow,tabularx}
\usepackage{balance}
\usepackage{longtable}
\usepackage{booktabs,longtable}
\usepackage{hhline}
\usepackage[flushleft]{threeparttable}
\usepackage{dsfont}
\usepackage{fancyhdr}
\usepackage{indentfirst}
\usepackage{epstopdf}
\usepackage{algorithm}
\usepackage{enumitem}
\usepackage{algorithmic}

\usepackage{makecell}

\newcolumntype{L}[1]{>{\raggedright\let\newline\\\arraybackslash\hspace{0pt}}m{#1}}
\newcolumntype{C}[1]{>{\centering\let\newline\\\arraybackslash\hspace{0pt}}m{#1}}
\newcolumntype{R}[1]{>{\raggedleft\let\newline\\\arraybackslash\hspace{0pt}}m{#1}}
\newfont{\mycrnotice}{ptmr8t at 7pt}
\newfont{\myconfname}{ptmri8t at 7pt}

\begin{document}
\title{Neural Compatibility Modeling with Attentive Knowledge Distillation}

\author{Xuemeng Song$^\dag$, Fuli Feng$^{\S}$, Xianjing Han$^\dag$, Xin Yang$^\dag$,  Wei Liu$^\sharp$, Liqiang Nie$^\dag$\textsuperscript{*}\\
        {$^{\dag}$Shandong University, $^{\S}$National University of Singapore}, $^{\sharp}$Tencent AI Lab\\
         \{sxmustc, fulifeng93, hanxianjing2018, joeyangbuer\}@gmail.com, wliu@ee.columbia.edu, nieliqiang@gmail.com\\
        }

\renewcommand{\shortauthors}{}

\begin{abstract}
\noindent Recently, the booming fashion sector and its huge potential benefits have attracted tremendous attention from many research communities. In particular, increasing research efforts have been dedicated to the complementary clothing matching as matching clothes to make a suitable outfit has become a daily headache for many people, especially those who do not
have the sense of aesthetics. Thanks to the remarkable success of neural networks in various applications such as image classification and speech recognition, the researchers are
enabled to adopt the data-driven learning methods to analyze fashion items. Nevertheless, existing studies overlook the rich valuable knowledge (rules) accumulated in fashion domain,
especially the rules regarding clothing matching. Towards this end, in this work, we shed light on complementary clothing matching by integrating the advanced deep neural networks
and the rich fashion domain knowledge. Considering that the rules can be fuzzy and different rules may have different confidence levels to different samples, we present a neural
compatibility modeling scheme with attentive knowledge distillation based on the teacher-student network scheme. Extensive experiments on the real-world dataset show the superiority of our model over several state-of-the-art baselines. Based upon the comparisons, we observe certain fashion insights that add value to the fashion matching study. As a byproduct, we released the codes, and involved parameters to benefit other researchers.
\end{abstract}

 \fancyhf{}

%
%



\maketitle
\section{Introduction}
\begin{figure}[!t]
\hspace*{0.0em} \subfigure[Composition1.]{\scalebox{0.21}{\includegraphics{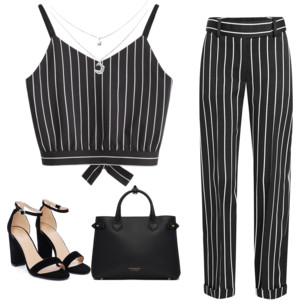}}\label{fig:subfig1}}\hspace*{1.8em}
 \subfigure[Composition2.]{\scalebox{0.155}{\includegraphics{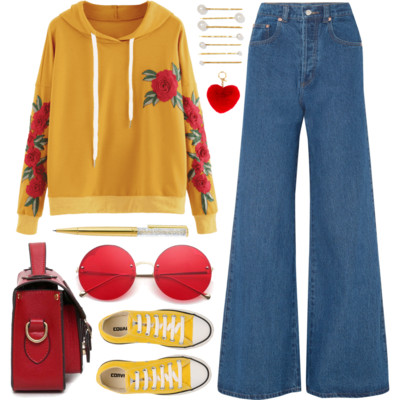}}\label{fig:subfig2}}\hspace*{1.2em}
 \subfigure[Composition3.]{\scalebox{0.21}{\includegraphics{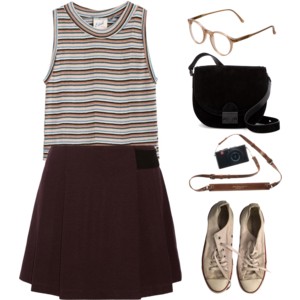}}\label{fig:subfig3}}
 \vspace{-1.2em}
\caption{Examples of outfit compositions.}
 \vspace{-1.6em}
\label{fig1}
\end{figure}

According to the Goldman Sachs, the $2016$ online retail market of China for fashion products, including apparel, footwear, and accessories,  has reached $187.5$ billion US dollars\footnote{\url{http://www.chinainternetwatch.com/19945/online-retail-2020}.}, which demonstrates people's great demand for clothing. In fact, clothing plays a pivotal role in people's daily life, as a proper outfit (e.g., a top with a bottom) can empower one's favorable impression. In a sense, how to make suitable outfits has become the daily headache of many people, especially those who do not have a good sense of clothing matching. Fortunately, recent years have witnessed the proliferation of many online fashion communities, such as Polyvore\footnote{\url{http://www.polyvore.com/}.} and Chictopia\footnote{\url{http://www.chictopia.com/}.}, where a great number of outfits composed by fashion experts have been made publicly available, as shown in Figure~\ref{fig1}. Based on such rich real-world data,  several researchers have attempted to intelligently aid people  in clothing matching.

In fact, most existing researches mainly rely on the deep neural networks to extract the effective representations for fashion items to tackle the clothing matching problem, due to their impressive advances in various research domains, including image classification, speech recognition and machine translation. However, as pure data-driven methods, neural networks not only suffer from the poor interpretability but also overlook the value of human knowledge.
Especially, as an essential aspect of people's daily life, clothing matching domain has accumulated various valuable knowledge, i.e., the matching rules. Although they may be of high subjectivity, certain matching rules have been widely accepted by the public as common sense. For example, tank tops would go better with shorts instead of dress, while silk tops better avoid knit bottoms. Therefore, it is highly desired to devise an effective model to seamlessly incorporate such domain knowledge into the pure data-driven learning methods and hence boost the performance.

%

\begin{figure}[!t]
  \centering
  \subfigure[Example1.]{
  \includegraphics[scale=0.53]{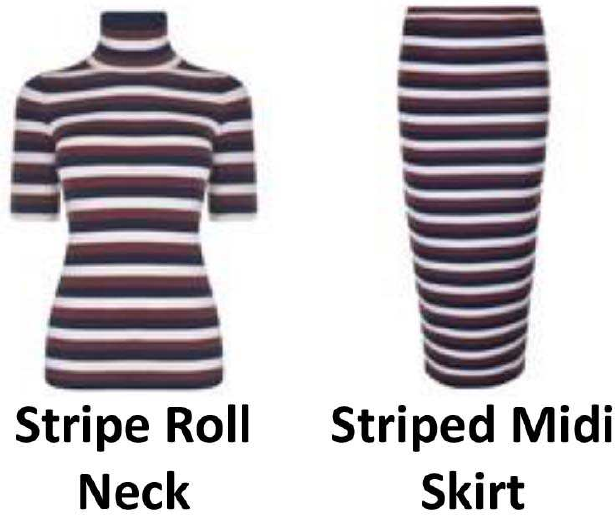}
   \label{fig:subfig1}
   }
   \subfigure[Example2.]{
  \includegraphics[scale=0.53]{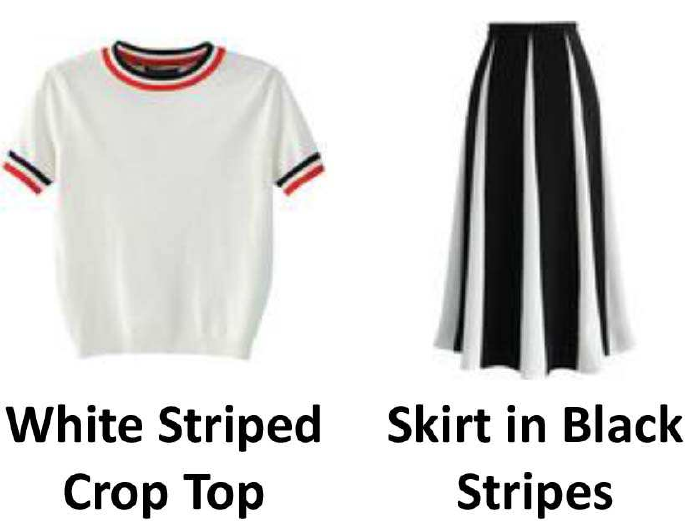}
   \label{fig:subfig1}
   }
    \vspace{-1.2em}
  \caption{Illustration of the rule confidence on different item pairs. Both examples satisfy the rule ``stripe tops can go with stripe bottoms''.}\label{fig2}
   \vspace{-1.6em}
\end{figure}

In this work, we aim to investigate the practical fashion problem of clothing matching by leveraging both the deep neural networks and the rich human knowledge in fashion domain. In fact, the problem we pose here can be cast as the compatibility modeling between the complementary fashion items, such as tops and bottoms. However, comprehensively model the compatibility between fashion items from both the data-driven and knowledge-driven perspectives is non-trivial due to the following challenges: 1) The human knowledge pertaining to fashion is usually implicitly conveyed by the compositions of fashion experts, which makes the domain knowledge unstructured and fuzzy. Therefore, how to construct a set of structured knowledge rules for clothing matching constitutes a tough challenge. 2) How to seamlessly encode such knowledge rules into the pure data-driven learning framework and enable the model to learn from
 not only the specific data but also the general rules poses another challenge for us. And 3) for different samples, knowledge rules may present different levels of confidence
 and hence provide different levels of guidance.
For example, as can be seen from Figure~\ref{fig2}, both compositions satisfy the rule ``stripe tops can go with stripe bottoms'' according to their contextual metadata. However, obviously, the given rule should impose more regularization towards the example of Figure~\ref{fig2}(a) and deserve higher rule confidence as compared to that of Figure~\ref{fig2}(b).  Accordingly, how to effectively assign the rule confidence is a crucial challenge.

\begin{figure*}[t]
\vspace{-0.8em}
  \centering
  \includegraphics[scale=0.76]{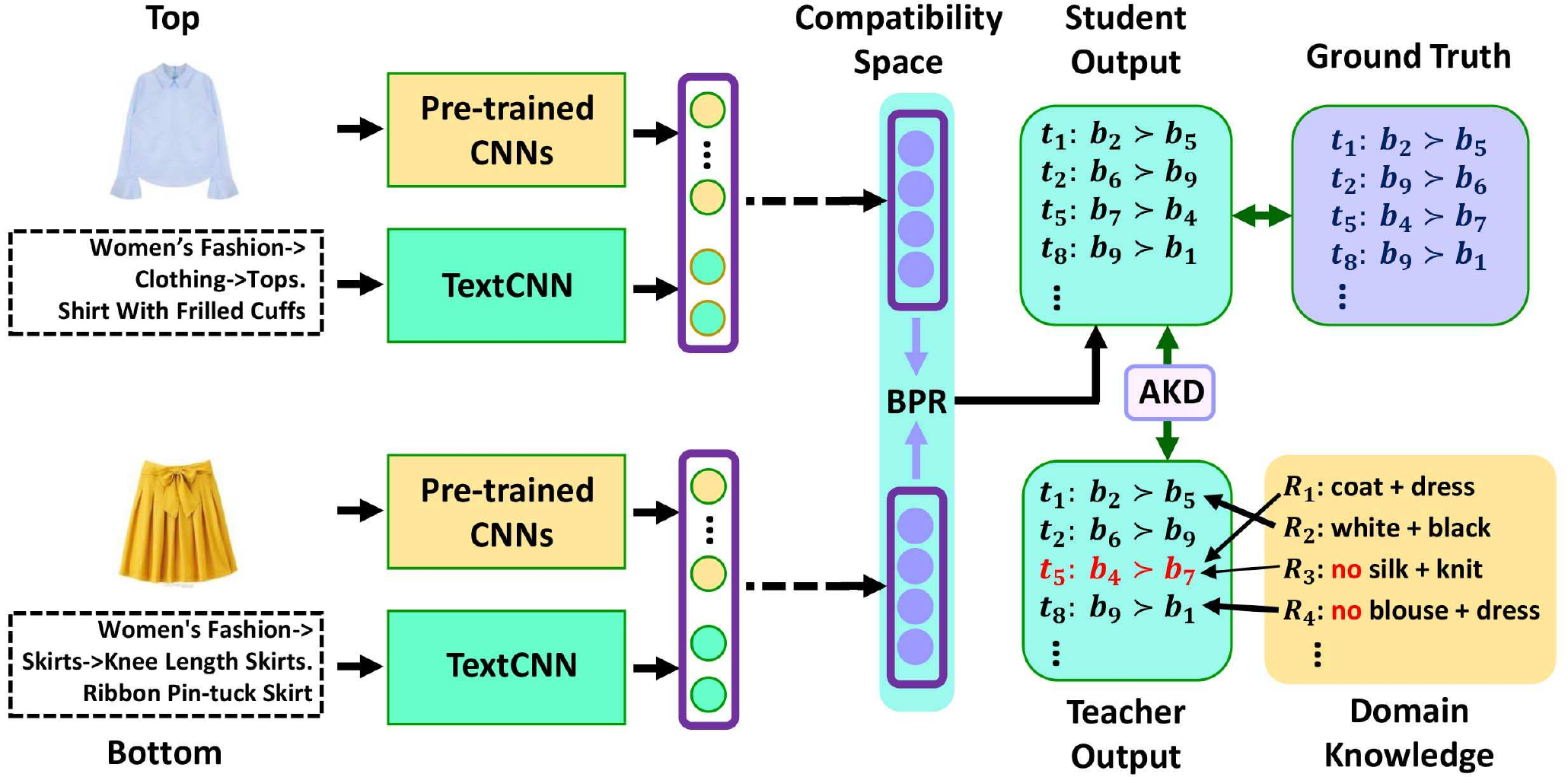}
  \caption{Illustration of the proposed scheme. The student network, consisting of dual-path neural networks, aims to
  learn the latent compatibility space where the implicit preference among items can be modeled via Bayesian Personalized Ranking.
  The teacher network encodes the domain knowledge and guides the student network
  via attentive knowledge distillation. $t_i$: top, $b_j$: bottom, ``$\succ$'': pair-wise preference. ``-$>$'' denotes the category hierarchy.  The width of the arrows originated from rules refers to the rule confidence.}\label{fig3}
    \vspace{-1em}
\end{figure*}

To address the aforementioned challenges, we present a compatibility modeling scheme with attentive knowledge distillation, dubbed as AKD-DBPR, as shown in Figure~\ref{fig3}, which is able to learn from both the specific data samples and the general domain knowledge.
In particular, we adopt the teacher-student scheme~\cite{HuMLHX16} to incorporate the domain knowledge (as a teacher) and enhance the performance of neural networks (as a student). As a pure data-driven learning, the student network aims to learn a latent compatibility space to unify the fashion items from heterogenous spaces with dual-path neural networks. To comprehensively model the compatibility and the semantic relation between different modalities, the student network seamlessly integrates the visual and contextual modalities of fashion items by imposing hidden layers over the concatenated vectors of visual and contextual representations. Moreover, to better characterize the
relative compatibility between fashion items, we investigate the pairwise preference between complementary fashion items by building our student network based on the Bayesian Personalized Ranking (BPR) framework~\cite{rendle2009bpr}. Meanwhile, we encode the domain knowledge with a set of flexible structured logic rules and encode these knowledge rules into the teacher network with regularizers, whereby we introduce the attention mechanism to attentively assign the rule confidence. Ultimately, the student network is encouraged to not only achieve good performance of the compatibility modeling
but also emulate the rule-regularized teacher network well.

Our main contributions can be summarized in threefold:

\begin{itemize}[leftmargin=2em]
\item
We present an attentive knowledge distillation scheme, which is able to encode the fashion
domain knowledge to the traditional neural networks.
To the best of our knowledge, this is the first to incorporate fashion domain knowledge to boost the compatibility modeling performance in the context of clothing matching.
\item Considering that different knowledge rules may have different confidence levels in the knowledge distillation procedure, we introduce
the attention mechanism to the proposed scheme to flexibly assign the rule confidence.
\item Extensive experiments conducted on the real-world dataset   demonstrate the superiority of the proposed scheme over the state-of-the-art methods.  As a byproduct, we released the codes, and involved parameters to benefit other researchers\footnote{\url{http://akd_dbpr.bitcron.com/}.}.
\end{itemize}

The remainder of this paper is structured as follows. Section $2$ briefly reviews the related work. The proposed AKD-DBPR is introduced in Section $3$. Section $4$ presents the experimental results and analyses, followed by our concluding remarks and future work in Section $5$.

\section{Related Work}
\subsection{Fashion Analyses}
Recently, the huge amount of potential benefits of fashion industry have attracted many researchers' attention from
 the computer vision to the multimedia research communities.
Existing efforts mainly focus on clothing retrieval~\cite{liu2012street,liu2012hi,hu2014style}, fashion trending prediction~\cite{GuWPS0K17}, fashionability prediction~\cite{li2017mining} and compatibility modeling~\cite{HanWJD17}.
For example, Liu et al.~\cite{liu2012hi} presented a latent Support Vector Machine~\cite{felzenszwalb2008discriminatively} model for both occasion-oriented outfit and item recommendation based on a dataset of wild street photos, constructed by manual annotations. Due to the infeasibility of human annotated dataset, several pioneering researchers have resorted to other sources, where real-world data can be harvested automatically. For example, Hu et al.~\cite{hu2015collaborative} investigated the problem of personalized outfit recommendation with a dataset collected from Polyvore. McAuley et al.~\cite{mcauley2015image} proposed a general framework to model the human visual preference for a given pair of objects based on the Amazon real-world co-purchase dataset. In particular, they extracted visual features with convolutional neural networks (CNNs) and introduced a similarity metric to model the human notion of complement objects. Similarly, He et al.~\cite{he2016vbpr} introduced a scalable matrix factorization approach that incorporates visual features of product images to fulfil the recommendation task. Although existing efforts have achieved compelling success, previous researches
 on fashion analysis mainly focus on the visual input but fail to consider the contextual information.
 Towards this end, several efforts have been dedicated to investigate the importance of the contextual
 information~\cite{li2017mining,SongFLLNM17} in fashion analysis. Overall, existing studies mainly focus on modeling the  compatibility purely based on the data-driven deep learning methods but overlook the value of domain knowledge.
 Distinguished from these researches, we aim to explore the potential of the fashion domain knowledge to  guide the pure data-driven neural networks and improve the interpretability as a side product.

\subsection{Knowledge Distillation}
Although deep neural networks have harvested huge success in a variety of application domains ranging from natural language processing to computer vision, several researchers still have
certain concerns with the poor interpretability as pure data-driven models. Towards this end, one mainstream research is to take advantage of the additional knowledge as a guidance to help to train the traditional neural networks. Hinton et al.~\cite{HintonVD15} first introduced a knowledge distillation framework to transfer the knowledge from a large cumbersome model to a small model. Inspired by this, Hu et al.~\cite{HuMLHX16} introduced an iterative teacher-student distillation approach, which combines neural networks with several first-order logic rules representing structured knowledge. Later, Yu et al.~\cite{YuLMD17} proposed to utilize the knowledge of linguistic statistics to regularize the learning process in the context of visual relationship detection. Although knowledge distillation in deep neural networks has been successfully applied to solve the visual relationship detection~\cite{YuLMD17}, sentence sentiment analysis and name entity recognition~\cite{rajendran2015bridge}, limited efforts have been dedicated to the fashion domain, which is the research gap we aim to bridge in this work.

\section{Neural Compatibility Modeling}

\subsection{Notation}
Formally, we first declare some notations. In particular, we use bold capital letters (e.g., $\mathbf{X}$) and bold lowercase letters (e.g., $\mathbf{x}$) to denote matrices and vectors, respectively. We employ non-bold letters (e.g., $x$) to represent scalars and Greek letters (e.g., $\beta$) to stand for parameters. If not clarified, all vectors are in column forms.  Let $\begin{Vmatrix}\mathbf{A}\end{Vmatrix}_F$  denote the Frobenius norm of matrix $\mathbf{A}$.

\subsection{Problem Formulation}
In a sense, people prefer to match compatible clothes to make a harmonious outfit. Accordingly, in this work, we aim to tackle the essential problem of compatibility modeling for clothing matching. Suppose we have a set of tops $\mathcal{T}=\{t_1, t_2, \cdots, t_{N_t}\}$ and bottoms $\mathcal{B}=\{b_1, b_2, \cdots, b_{N_b}\}$, where $N_t$ and $N_b$ denote the total numbers of tops and bottoms, respectively. For each $t_i$ ($b_i$), we use $\mathbf{v}_i^t$ ($\mathbf{v}_i^b$) $\in \mathbb{R}^{D_v}$ and $\mathbf{c}_i^t$ ($\mathbf{c}_i^b$) $\in \mathbb{R}^{D_c}$ to represent its visual and contextual embeddings, respectively. $D_v$ and $D_c$ denote the dimensions of the corresponding embeddings. In addition,  we have a set of positive top-bottom pairs $\mathcal{S}=\{(t_{i_1}, b_{j_1}), (t_{i_2}, b_{j_2}), \cdots, (t_{i_N}, b_{j_N})\}$ derived from the set of outfits composed by fashion experts on Polyvore, where $N$ is the total number of positive pairs. Accordingly, for each top $t_i$, we can derive a set of positive bottoms $\mathcal{B}_{i}^{+}=\{b_j\in \mathcal{B} | (t_i, b_j) \in \mathcal{S}\}$. Meanwhile, we have a set of rules $\mathcal{R}=\{R_l\}_{l=1}^L$ pertaining to clothing matching, where $R_l$ is the $l$-th rule and $L$ is the total number of rules. We employ $\mathcal{L}^+$ and $\mathcal{L}^-$ denote the set of positive and negative rules, respectively.  Let $m_{ij}$ denote the compatibility between top $t_i$ and bottom $b_j$, and on top of that we can generate a ranking list of bottoms $b_j$'s for a given top $t_i$ and hence solve the practical problem of clothing matching.
To accurately measure $m_{ij}$, we focus on devising a neural compatibility modeling scheme, which is able to jointly learn from both the specific data samples and general knowledge rules.

\subsection{Data-driven Compatibility Modeling}
Apparently, it is not advisable to directly measure the compatibility between complementary fashion items from the original distinct spaces due to their heterogeneity. Similar to~\cite{SongFLLNM17}, we assume that there is a latent compatibility space that can bridge the gap between the fashion items from the heterogenous spaces. In such latent space, compatible complementary fashion items are enabled to share high similarity. Considering that the factors contributing to the compatibility between fashion items may diversely range from style and color, to material and shape, and their relations can be rather sophisticated,
 we assume that the compatibility space is highly non-linear. In particular,  we adopt the pure data-driven neural network to explore the latent compatibility space, due to its recent compelling success in various machine learning applications.

As a matter of fact, each fashion item can be associated with multiple modalities, such as visual and contextual, and different modalities complementarily characterize the same fashion item. In particular, the visual modality can intuitively reflect the color and shape of the fashion items, while the contextual modality can briefly summarize the category and material information. To seamlessly exploit the potential of both modalities in the compatibility modeling, we employ the multi-layer perceptron (MLP) to model the semantic relation between different modalities of the same fashion items. In particular, we add $K$ hidden layers over the concatenated vectors of visual and contextual representations as follows,
\begin{equation}
\left \{
\begin{aligned}
&\mathbf{z}_{i0}^x= \begin{bmatrix}\mathbf{v}_i^x\\ \mathbf{c}_i^x\end{bmatrix}, \\
&\mathbf{z}_{i1}^x=s(\mathbf{W}_1^x\mathbf{z}_{i0}^x+\mathbf{b}_1^x), \\
&\mathbf{z}_{ik}^x=s(\mathbf{W}_k^x\mathbf{z}_{i(k-1)}^x+\mathbf{b}_{k}^x),\  k=2, \cdots, K, x=\{t, b\},
\end{aligned}
\right.
\end{equation}
where $\mathbf{z}_{ik}^x$ denotes the hidden representation, $\mathbf{W}_k^x$ and $\mathbf{b}_k^x$, $k=1, \cdots, K$, are weight matrices and biases, respectively. The superscripts $t$ and $b$ refer to \emph{top} and \emph{bottom}. $s:\mathbb{R}\mapsto \mathbb{R}$ is a non-linear function applied element wise\footnote{In this work, we use the sigmoid function $s(x)=1/(1+e^{-x})$.}.
We treat the output of the $K$-th layer as the latent representations for tops and bottoms, i.e., $\tilde{\mathbf{z}}_i^x=\mathbf{z}_{iK}^x\in \mathbb{R}^{D_l}, x=\{t, b\}$, where $D_l$ denotes the dimensionality of the latent compatibility space. Accordingly, we can measure the compatibility between top $t_i$ and bottom $b_j$ as follows,
\begin{align}\label{eq2}
m_{ij}&=(\tilde{\mathbf{z}}_i^t)^{T}\tilde{\mathbf{z}}_j^b.
\end{align}

In a sense, we can easily derive the positive (compatible) top-bottom pairs from those have been composed together by fashion experts. However, pertaining to the non-composed fashion item pairs, we cannot draw the conclusion that they are incompatible as they can also be the missing potential positive pairs (i.e., pairs can be composed in the future). Towards this end, to accurately model the implicit relations between the tops and bottoms, we naturally adopt the BPR framework, which has proven to be effective in the implicit preference modeling~\cite{He2016Fast}. In particular, we assume that bottoms from the positive set $\mathcal{B}_{i}^{+}$ are more compatible to top $t_i$ than those non-composed neutral bottoms. Accordingly, we build the following training set:
\begin{align}\label{eq4}
\mathcal{D}_{S}:=\{(i,j,k)| t_i \in \mathcal{T}, b_j\in\mathcal{B}_{i}^{+} \wedge b_k\in \mathcal{B} \backslash \mathcal{B}_{i}^{+} \},
\end{align}
where the triplet $(i,j,k)$ indicates that bottom $b_j$ is more compatible with top $t_i$ compared to bottom $b_k$.

Then according to~\cite{rendle2009bpr}, we have the objective function,
\begin{align}\label{eq5}
\mathcal{L}_{bpr}&=\sum_{(i,j,k)\in \mathcal{D}_S}\mathcal{L}_{bpr}(m_{ij},m_{ik})\nonumber \\
&=\sum_{(i,j,k)\in \mathcal{D}_S}-ln(\sigma(m_{ij}-m_{ik}))+\frac{\lambda}{2}\begin{Vmatrix}\Theta\end{Vmatrix}_F^2,
\end{align}
where $\lambda$ is the non-negative hyperparameter, the last term is designed to avoid overfitting and $\Theta$ refers to the set of parameters (i.e., $\mathbf{W}_k^x$ and $\mathbf{b}_{k}^x$) of neural networks.

\subsection{Attentive Knowledge Distillation}
As an important aspect of people's daily life, clothing matching has gradually accumulated much valuable human knowledge.
 For example, it is favorable that a coat goes better with a dress than with a short pants, while a silk top can hardly go with a knit bottom. In order to fully leverage the valuable domain knowledge, we utilize the knowledge distillation technique to guide the neural networks and allow the model to learn from general rules~\cite{HuMLHX16}. In particular, we adopt the teacher-student scheme, whose underlying intuition is analogous to the human education, where the teacher is aware of several professional rules and he/she thus can instruct students with his/her solutions to particular questions. In this work, considering the flexibility of logic rules as a declarative language, we use logic rules to represent the fashion domain knowledge. We encode these rules via regularization terms into a teacher network $q$, which can be further employed to guide the training of the student network $p$ of interest (i.e., the aforementioned data-driven neural network designed for  compatibility modeling). Ultimately, we aim to achieve a good balance between the superior prediction performance of student network $p$ and the mimic capability of student network $p$ to teacher network $q$.  Accordingly, we have the objective formulation at iteration $t$ as,
\begin{align}\label{eq6}
\boldsymbol{\Theta}^{(t+1)}=&\arg\min_{\boldsymbol{\Theta}} \sum_{(i,j,k)\in \mathcal{D}_S}\Big\{(1-\rho)\mathcal{L}_{bpr}(m_{ij}^p,m_{ik}^p)\nonumber \\
&+\rho\mathcal{L}_{crs}\Big(\mathbf{q}^{(t)}(i,j,k),\mathbf{p}(i,j,k)\Big)\Big\}+\frac{\lambda}{2}\begin{Vmatrix}\Theta\end{Vmatrix}_F^2,
\end{align}
where $\mathcal{L}_{crs}$ stands for the cross-entropy loss, $\mathbf{p}(i,j,k)$ and $\mathbf{q}(i,j,k)$ refer to the sum-normalized distribution over the compatibility scores predicted by the student network $p$ and teacher network $q$, (i.e., $[m_{ij}^p, m_{ik}^p]$ and $[m_{ij}^q, m_{ik}^q]$), respectively. $\rho$ is the imitation parameter calibrating the relative importance of these two objectives.

\subsection{Teacher Network Construction}

\begin{figure}[!t]
\vspace{-0.8em}
  \centering
  \includegraphics[scale=0.58]{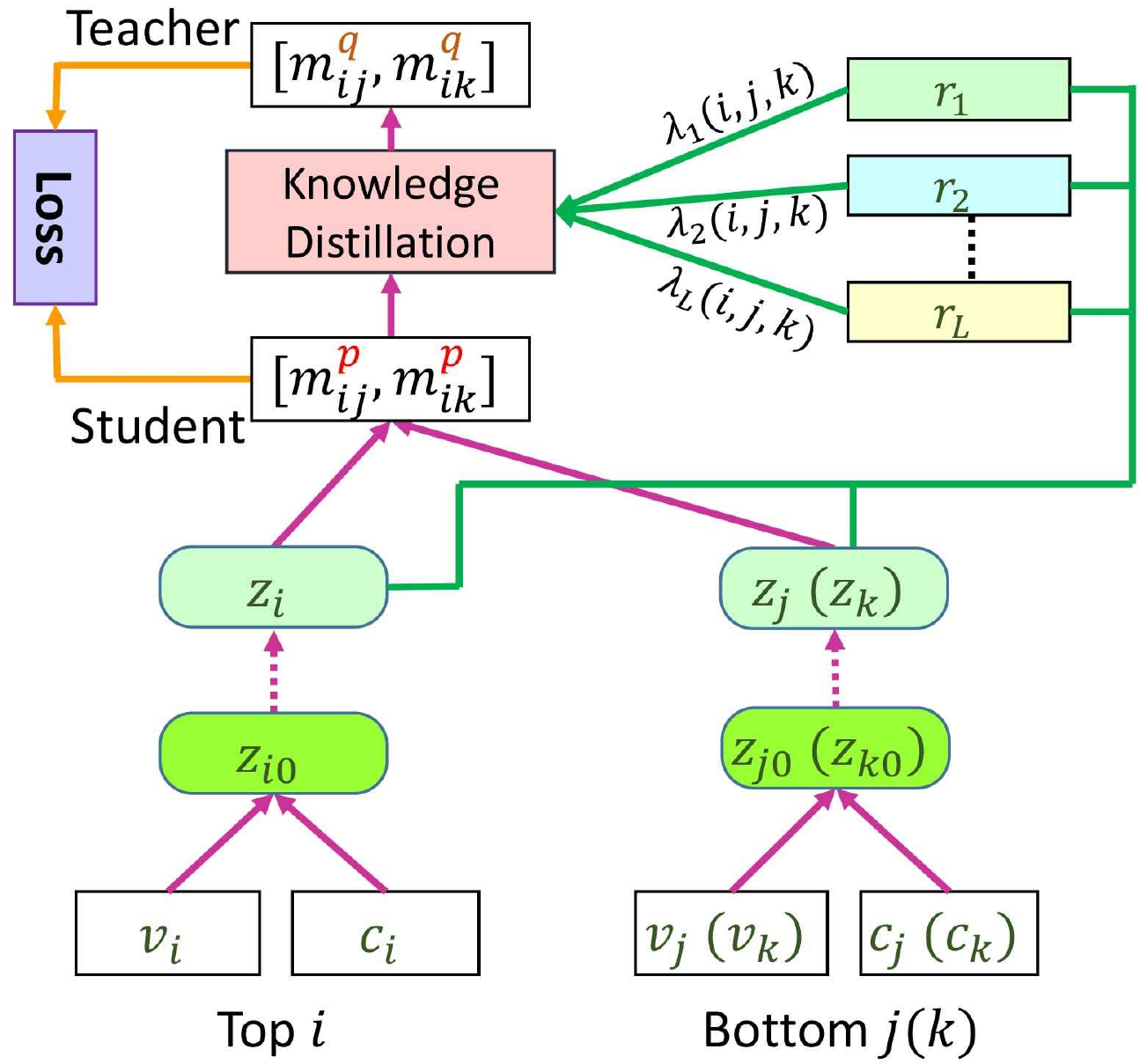}
  \vspace{-1em}
  \caption{Workflow of the proposed attentive knowledge distillation framework.}\label{fig44}
    \vspace{-1.5em}
\end{figure}

As the teacher network plays a pivotal role in the knowledge distillation process, we now proceed to introduce the derivation of the teacher network $q$. On the one hand, we expect that the student network $p$ can learn well from the teacher network $q$ and such property can be naturally measured by the closeness between the compatibility prediction of both networks $p$ and $q$. On the other hand, we attempt to utilize the rule regularizer to encode the general domain knowledge. In particular, we adapt the teacher network construction method proposed in~\cite{HuMLHX16,HuYSX16} as follows,
\begin{align}\label{eq9}
\min_{\mathbf{q}}& KL(\mathbf{q}(i,j,k) \parallel \mathbf{p}(i,j,k))-C\sum_{l}\mathbb{E}_{\mathbf{q}}[\mathbf{f}_{l}(i,j,k)],
\end{align}
where $C$ is the balance regularization parameter and $KL$ measures the KL-divergence between $\mathbf{p}(i,j,k)$ and $\mathbf{q}(i,j,k)$. This formulation has proven to be a convex problem and can be optimized with the following closed-form solutions,
\begin{align}\label{eq10}
 \mathbf{q}^*(i, j, k) \propto   \mathbf{p}(i, j, k)exp\Big\{\sum_{l}C \lambda_l\mathbf{f}_{l}(i, j, k)\Big\},
\end{align}
where $\lambda_l$ stands for the confidence of the $l$-th rule and the larger $\lambda_l$ indicates the stronger rule constraint.  $\mathbf{f}_{l}(i, j, k)$ is the $l$-th rule constraint function devised to reward the predictions of the student network that meet the rules while penalize the others. In our work, given the sample $(i,j,k)$, we expect to reward the compatibility  $m_{ij}$, if $(i,j)$ satisfies the positive rule but $(i,k)$ not or $(i,k)$ triggers the negative rule while $(i,j)$ not. In particular, we define  $f_{l}^{ij}(i, j, k)$, the element of $\mathbf{f}_{l}(i, j, k)$ for $m_{ij}$, as follows,
\begin{align}\label{eq11}
 f_{l}^{ij}(i, j, k)=
\begin{cases}
   1, & \text{if} \begin{cases}\delta_l(ij)=1, \delta_l(ik)=0, l\in \mathcal{L}^{+},\\ \delta_l(ij)=0, \delta_l(ik)=1, l\in \mathcal{L}^{-},\end{cases}\\
   0, & \text{others,}
\end{cases}
\end{align}
where $\delta_l(ab)=1 (0)$ means that the sample $(a,b)$ satisfies the $l$-th rule (or not). We define the other element  $f_l^{ik}(i,j,k)$ of $\mathbf{f}_{l}(i, j, k)$ regarding $m_{ik}$ in the same manner.


\begin{algorithm}[!t]
\caption{Attentive Knowledge Distillation.} \label{alg1}
\begin{algorithmic}[1]
\REQUIRE ~~\
$\mathcal{D}_S=\{(i,j,k)\}$, $\mathcal{R}=\{(R_l)\}_{l=1}^L$,  $\rho$, $C$\
\ENSURE ~~\
Parameters $\boldsymbol{\Theta}$ in the student network $p$, parameters $\boldsymbol{\Phi}$ in the attention network $a$.
\STATE Initialize neural network parameters $\boldsymbol{\Theta}$ and $\boldsymbol{\Phi}$.
\REPEAT
\STATE Draw $(i,j,k)$ from $\mathcal{D}_S$
\FOR{each $l$ in $\mathcal{L}(i,j,k)$}
\STATE Compute $\lambda_l(i,j,k)$ according to Eqns.~($\ref{eq12}$) and ($\ref{eq13}$).
\ENDFOR
\STATE Construct teacher network $q$ according to Eqn.~($\ref{eq10}$).
\STATE Transfer knowledge into $p$ by updating $\boldsymbol{\Theta}$ and $\boldsymbol{\Phi}$ according to Eqn.~($\ref{eq6}$).
\UNTIL {Converge}
\end{algorithmic}
\end{algorithm}

Traditionally, $\lambda_l$ in Eqn.($\ref{eq10}$) can be either manually assigned or automatically learnt from the data, and both ways assume the rules have universal confidence to all samples. However, in fact, different rules may have different confidence levels for different samples, which can be attributed to the fact that the human knowledge rules can be general and fuzzy. It is intractable to directly
pre-define the universal rule confidence. Therefore, considering that different rules can flexibly contribute to the guidance to the given samples, we adopt the attention mechanism, which has proven to be effective in many machine learning tasks such as multimedia recommendation~\cite{ChenZ0NLC17}, representation learning~\cite{QuTSR0017} and extractive summarization~\cite{RenCRWMR17}. The key to the success of attention mechanism lies in the observation that human tends to selectively attend to parts of the input signal rather than the entirety  at once during the process of human recognition. In our work, we adopt the soft attention model to assign the rule confidence adaptively according to the given samples.  In particular, for a given sample $(i,j,k)$ and the set of rules it activates $\mathcal{L}(i,j,k)$, we assign $\lambda_l(i,j,k)$  as follows,
\begin{align}\label{eq12}
 \lambda_l^'(i,j,k)=&\mathbf{w}^T\phi(\mathbf{W}_t[\tilde{\mathbf{v}}_i,\tilde{\mathbf{t}}_i]+\mathbf{W}_b[\tilde{\mathbf{v}}_j,\tilde{\mathbf{t}}_j]+\mathbf{W}_b[\tilde{\mathbf{v}}_k,\tilde{\mathbf{t}}_k]\nonumber \\
 &+\mathbf{W}_r\mathbf{r}_l+\mathbf{b})+c, \text{\ \ } l \in \mathcal{L}(i,j,k),
\end{align}
where the $\mathbf{W}_t\in \mathbb{R}^{h\times(D_v+D_t)}$, $\mathbf{W}_b\in \mathbb{R}^{h\times(D_v+D_t)}$, $\mathbf{W}_l\in \mathbb{R}^{h\times L}$,  $\mathbf{w}\in \mathbb{R}^{h}$, $\mathbf{b}\in \mathbb{R}^{h}$ and $c$ are the model parameters. $h$ represents the hidden layer size of the attention network. $\mathbf{r}_l\in \mathbb{R}^{L}$ stands for the one-hot encoding of the $l$-th rule. The attention scores are then normalized as follows,
\begin{align}\label{eq13}
 \lambda_l(i,j,k)=\frac{exp(\lambda_l^'(i,j,k))}{\sum_{u\in \mathcal{L}(i,j,k)}exp(\lambda_{u}^'(i,j,k))}.
\end{align}

Figure~\ref{fig44} illustrates the workflow of our model, while the optimization procedure of our framework is summarized in Algorithm~\ref{alg1}. Notably, the teacher network is first constructed from the student network at the very beginning of the training, which may induce the poor guidance at first. Therefore, we expect the whole framework favors to the prediction of the ground truth more at the initial stage but gradually bias towards the imitate capability of the student network to the teacher network. Therefore, we adopt the parameter assigning strategy in~\cite{HuMLHX16} to assign
 $\rho$ dynamically, which keeps $\rho$  increasing as the training process goes.

\subsection{Rules Construction}
In this work, we aim to leverage the explicit structured domain knowledge to guide the student neural network and hence boost the performance.  To derive the domain knowledge, we first exploit our internal training dataset, which contains rich positive  top-bottom pairs. In general, the compatibility between fashion items mainly affected by five attributes: color, material, pattern, category and brand. We hence define a dictionary with the possible values of each attribute based on the training dataset while taking the  annotation details in~\cite{MaJZFLT17} as a reference. Due to the limited space, Table~\ref{table9} shows several value examples of each attribute\footnote{The complete list can be accessed via \url{http://akd_dbpr.bitcron.com/}.}. We then calculate the co-occurrence of the value pairs for each attribute and retain both the top $10$ and the last $10$ pairs as the rule candidates, as we assume that the high co-occurrence can indicate the high compatibility and facilitate the screen of positive rules, e.g., ``black top goes better with a black bottom'', while the low co-occurrence may contribute to the derivation of the negative rules, such as ``blouse cannot go with the dress''. The underlying philosophy behind is that sometimes it is intractable to identify compatible fashion items but effortless to determine the incompatible ones. Thereafter, to ensure the quality of these rules extracted from the limited dataset, we further ask three fashion-lovers to manually screen the final rules.  Finally, we obtain $15$ rules, which we will discuss in detail in the following section.

\begin{table}[!t]
\centering \caption{Value examples of each attribute.}\label{table9}
\vspace{-1em}
  \begin{tabular}{|C{1.5cm}||*{1}{C{6cm}|}}
     \hline
     Attribute & Value Examples \\\hline \hline
     Color & black, white, green, red, blue, grey \\\hline
     Material& knit, silk, leather, cotton, fur, cashmere\\\hline
     Pattern& pure, grid, dot, floral, number (letter)  \\\hline
     Category& coat, dress, skirt, sweater, jeans, hoodie\\\hline
     Brand& Yoins, HM, Topshop, Gucci\\\hline
   \end{tabular}
   \vspace{-1em}
\end{table}
For simplicity, we use ``\textit{value}$1$ + \textit{value}$2$'' to denote the positive rule, while ``no \textit{value}$1$ + \textit{value}$2$'' representing the negative rule. For example, ``black + black'' stands for the positive rule ``black tops can go with black bottoms'', and ``no silk + knit'' represents the negative rule ``silk tops cannot go with knit bottoms''. 
According to  Eqn.($\ref{eq11}$), our model needs to determine whether the pair of $t_i$ and $b_j$ activates the given rule.  We hence argue that $(t_i, b_j)$ satisfies the (positive/negative) rule, if the \textit{value}$1$ and \textit{value}$2$ of the rule respectively appear in the contextual metadata of $t_i$ and $b_j$.

\section{Experiment}
To evaluate the proposed method, we conducted extensive experiments on the real-world dataset \textbf{FashionVC} by answering the following research questions:
\begin{itemize}[leftmargin=7pt]
  \item Does AKD-DBPR outperform the state-of-the-art methods?
  \item How do the attention mechanisms affect the performance?
  \item How do AKD-DBPR perform in the application of the complementary fashion item retrieval?
\end{itemize}

\subsection{Experimental Settings}

\textbf{Dataset.} In this work, we adopted the publicly released dataset  \textbf{FashionVC}~\cite{SongFLLNM17} to evaluate our proposed model. FashionVC consists of $20,726$ outfits with $14,871$ tops and $13,663$  bottoms, composed by the fashion experts on Polyvore. Each fashion item in FashionVC is associated with a visual image, relevant categories and the title description.

\textbf{Contextual Representation.}
In this work, contextual description of each fashion item refers to its title and category labels in different granularity. To obtain the effective contextual representation, we adopted the CNN architecture~\cite{Kim14}, which has achieved compelling performance in various natural language processing tasks~\cite{SeverynM15a}. In particular, we first represented each contextual description as a concatenated word vector, where each row represents one constituent word and each word is represented by the publicly available $300$-D word2vec vector. We then deployed  the single channel CNN, consisting of a convolutional layer on top of the concatenated word vectors and a max pooling layer. In particular, we have four kernels with sizes of $2$, $3$, $4$, and  $5$, $100$ feature maps for each and the rectified linear unit (ReLU) as the activation function. Ultimately, we obtained a $400$-D contextual representation for each item.

\textbf{Visual Representation.}
Regarding the visual modality, we applied the deep CNNs, which has proven to be the state-of-the-art model for image representation learning~\cite{chen2016micro,khosla2014makes,mcauley2015image}. In particular, we chose the pre-trained ImageNet deep neural network provided by the Caffe software package~\cite{jia2014caffe}, which consists of $5$ convolutional layers followed by $3$ fully-connected layers. We fed the image of each fashion item to the CNNs, and adopted the fc7 layer output as the visual representation. Thereby, we represented the visual modality of each item with a $4096$-D vector.
¡¡

We divided the positive pair set $\mathcal{S}$ into three chunks: $80\%$ of triplets for training, $10\%$ for validation, and $10\%$ for testing, denoted as $\mathcal{S}_{train}$, $\mathcal{S}_{valid}$ and $\mathcal{S}_{test}$, respectively. We then generated the triplets $\mathcal{D}_{S_{train}}$, $\mathcal{D}_{S_{valid}}$ and $\mathcal{D}_{S_{test}}$ according to Eqn.($\ref{eq4}$). For each positive pair of $t_i$ and $b_j$, we randomly sampled $M$ bottoms $b_k$'s and each $b_k$ contributes to a triplet $(i,j,k)$, where  $b_k \notin \mathcal{B}_i^+$ and $M$ is set as $3$. We adopted the area under the ROC curve (AUC)~\cite{rendle2010pairwise,zhang2013attribute} as the evaluation metric. For optimization, we employed the stochastic gradient descent (SGD)~\cite{bottou1991stochastic} with the momentum factor as $0.9$. We adopted the grid search strategy to determine the optimal values for the regularization parameters (i.e., $\lambda, C$) among the values $\{10^{r}| r \in \{-4,\cdots, -1\}\}$ and $[2, 4, 6, 8]$, respectively. In addition, the mini-batch size, the number of hidden units and learning rate  were searched in $[32, 64, 128, 256]$, $[128, 256, 512, 1024]$, and $[0.01, 0.05, 0.1]$, respectively. The proposed model was fine-tuned for $40$ epochs, and the performance on the testing set was reported. We empirically found that the proposed model achieves the optimal performance with $K=1$ hidden layer of $1024$ hidden units.

\begin{figure}[!t]
  \centering
  \subfigure[]{
  \includegraphics[scale=0.225]{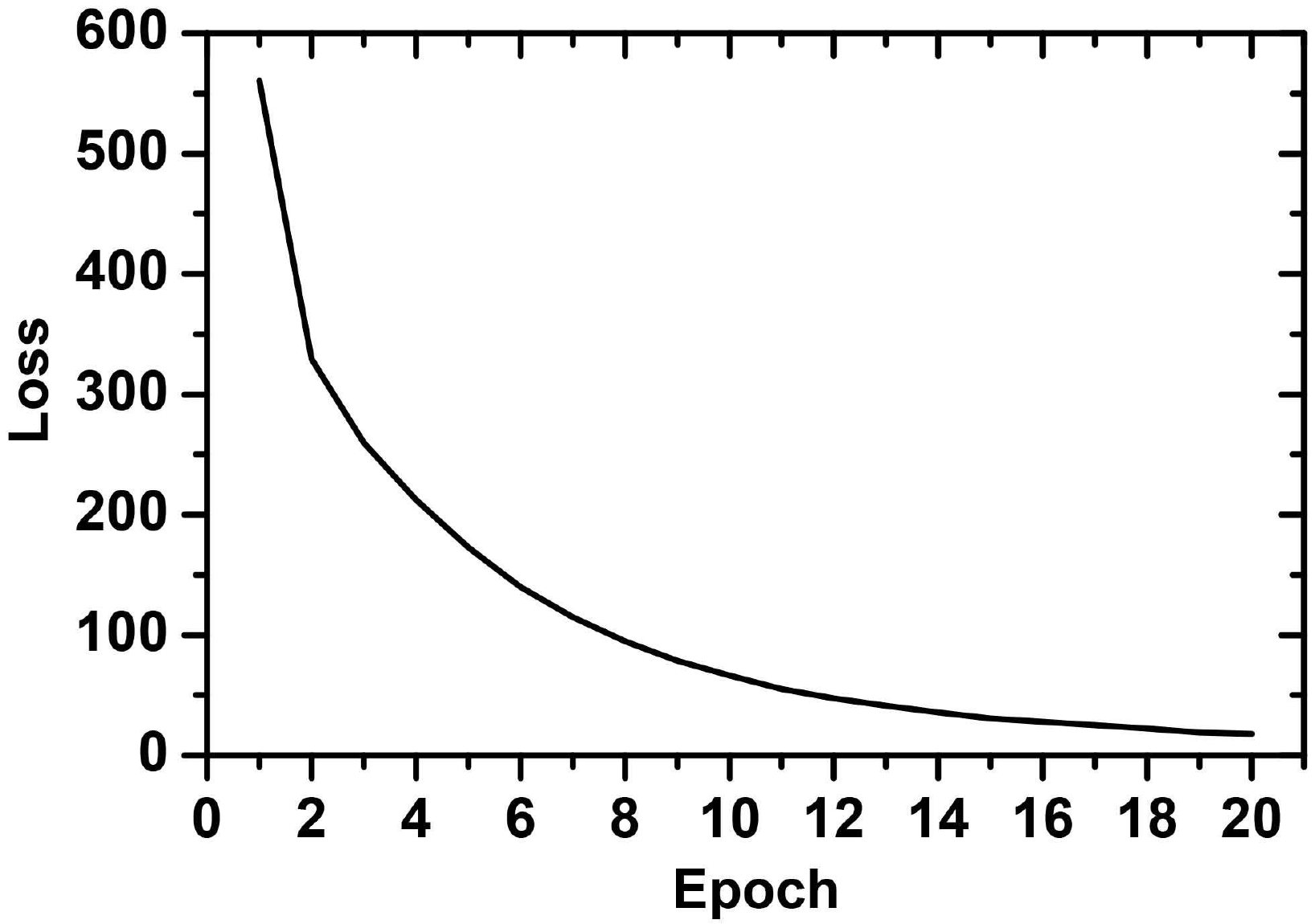}
   \label{fig:subfig1}
   }
   \subfigure[]{
  \includegraphics[scale=0.225]{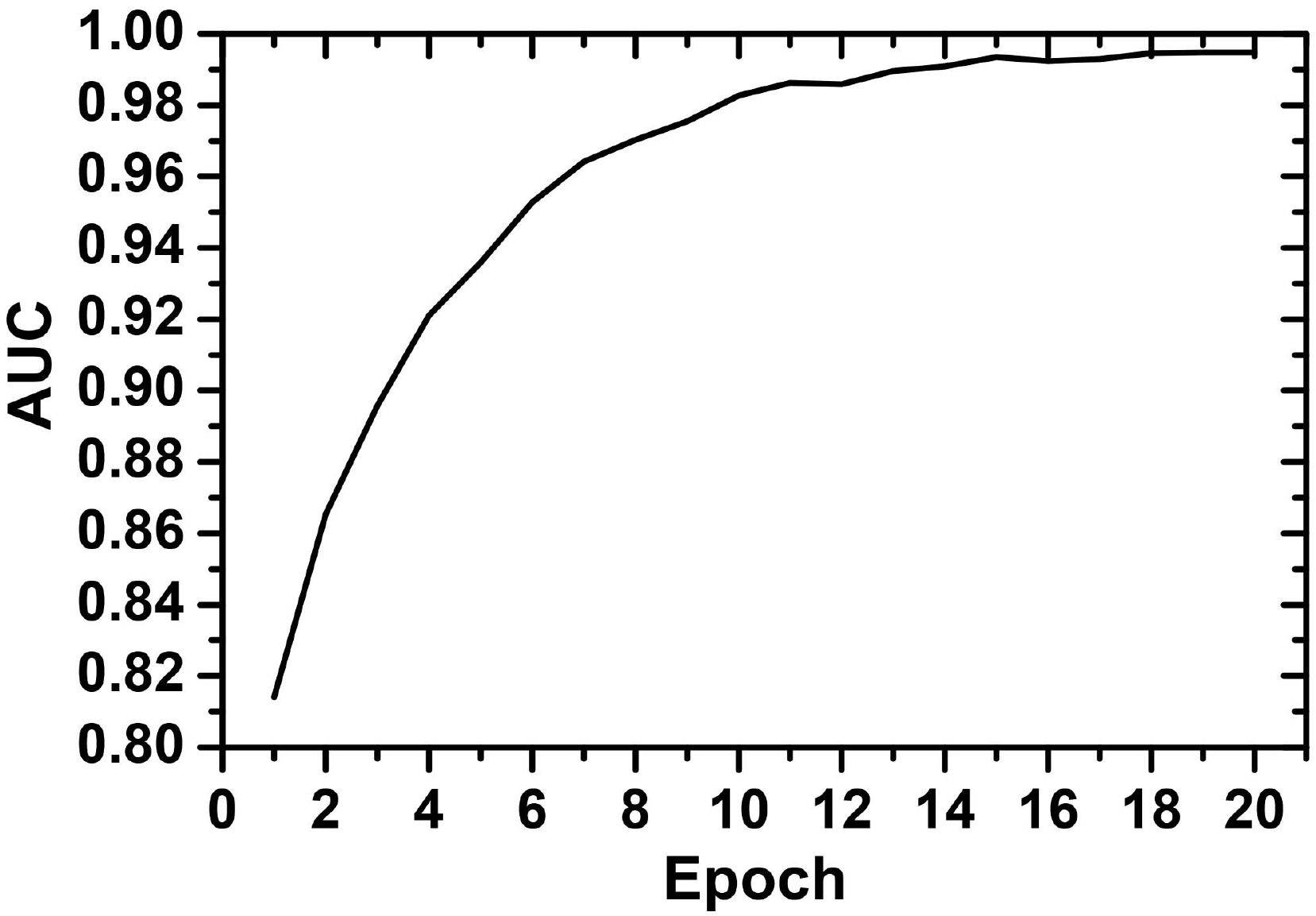}
   \label{fig:subfig1}
   }
    \vspace{-1.2em}
  \caption{Training loss and the AUC curves.}\label{fig4}
   \vspace{-1.2em}
\end{figure}

We first experimentally verified the convergence of the proposed learning scheme. Figure~\ref{fig4} shows the changes of the objective function in Eqn.($\ref{eq6}$) and the training AUC with one iteration of our algorithm. As we can see, both values first change rapidly in a few epochs and then go steady finally, which well demonstrates the convergence of our model.
\begin{figure*}[!b]
  \centering
    \vspace{-0.8em}
  \includegraphics[scale=0.585]{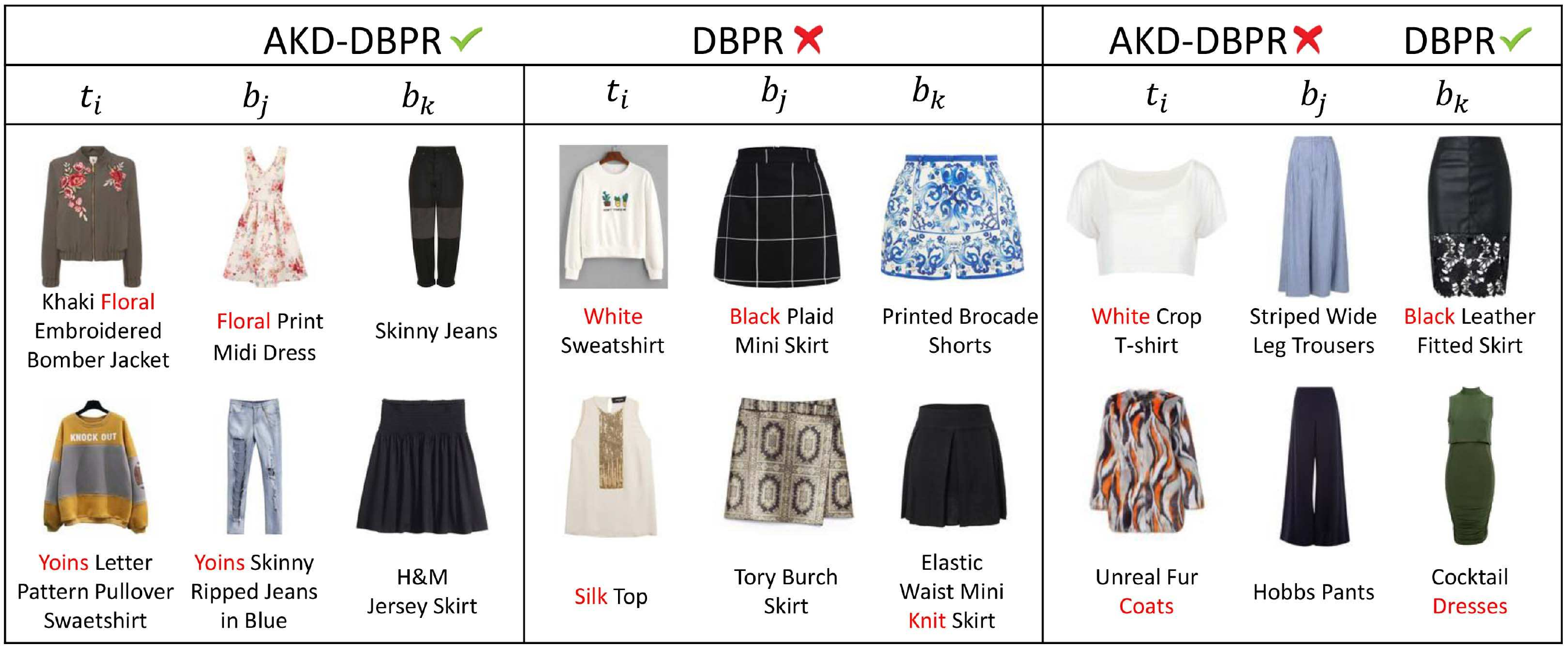}
  \vspace{-0.8em}
  \caption{Comparison between AKD-DBPR and DBPR on testing triplets. All the triplets satisfy that $t_i: b_j \succ b_k$. We only list the keywords of the metadata of items and highlight the values of the rule. }\label{fig5}
    \vspace{-1.2em}
\end{figure*}
\subsection{On Model Comparison (RQ1)}
Due to the sparsity of our dataset, where matrix factorization based methods~\cite{wang2014improving,qian2014personalized} are not applicable, we chose the following content-based baselines regarding compatibility modeling to evaluate the proposed model AKD-DBPR.
\begin{itemize}[leftmargin=15pt]
\item \textbf{POP}: We used the  ``popularity'' of bottom $b_j$ to measure its compatibility with top $t_i$. Here the ``popularity'' is defined as the number of tops that has been paired with $b_j$ in the training set.
\item \textbf{RAND}: We randomly assigned the compatibility scores of $m_{ij}$ and $m_{ik}$ between items.

\item \textbf{IBR}: We chose the image-based recommendation method proposed by~\cite{mcauley2015image}, which aims to model the compatibility between objects based on their visual appearance. This method learns a latent style space, where the retrieval of related objects can be performed by traditional nearest-neighbor search. Different from our model, this baseline learns the latent space by simple linear transformation and only consider the visual information of fashion items.

\item  \textbf{ExIBR}: We adopted the extension of IBR in~\cite{SongFLLNM17}, which is able to handle both the visual and contextual data of fashion items.

\item \textbf{BPR-DAE}: We selected the content-based neural scheme introduced by~\cite{SongFLLNM17}, which is capable of jointly modeling the coherent relation between different modalities of fashion items and the implicit preference among items via a dual autoencoder network.

\item \textbf{DBPR}: To get a better understanding of our model,  we introduced the baseline DBPR, which is the derivation of our model by removing the guidance of the teacher network and solely relies on the student network.
\end{itemize}

 \begin{table}[!t]
\centering \caption{Performance comparison among different approaches in terms of AUC.}\label{table6}
\vspace{-1em}
  \begin{tabular}{|L{3cm}||*{1}{C{2.1cm}|}}
     \hline
     Approaches & AUC \\\hline \hline
     \textbf{POP}& $0.4206$\\
     \textbf{RAND}& $0.5094$ \\
     \textbf{IBR}& $0.6075$\\
     \textbf{ExIBR}& $0.7033$\\
     \textbf{BPR-DAE}&$0.7616$ \\\hline
     \textbf{DBPR}&$0.7704$ \\
     \textbf{AKD-DBPR-p}&$0.7843$ \\
     \textbf{AKD-DBPR-q}&$0.7852$ \\\hline
   \end{tabular}
   \vspace{-1.5em}
\end{table}

Since we can choose either the distilled student network $p$ or the teacher network $q$ with a final projection according to Eqn.($\ref{eq10}$) for the testing, we introduced two derivations of our model:
AKD-DBPR-p and AKD-DBPR-q. Here $p$ ($q$) means to use the final student (teacher) network to calculate the compatibility between items according to Eqn.($\ref{eq2}$).

Table~\ref{table6} shows the performance comparison among different approaches. From this table, we have the following
observations: 1) DBPR outperforms all the other state-of-the-art pure data-driven baselines, which indicates the superiority of the proposed content-based neural networks for compatibility modeling. 2) AKD-DBPR-p and AKD-DBPR-q both surpass DBPR, which validates the benefit of knowledge distillation in the context of compatibility modeling.
To intuitively understand the impact of the rule guidance, we illustrate the comparison between AKD-DBPR and DBPR on several testing triplets in Figure~\ref{fig5}. As we can see, AKD-DBPR performs especially better in cases when the given two bottoms $b_j$ and $b_k$ both seem to be visually compatible to the top $t_i$. Nevertheless, the general knowledge rules may also lead to the failed triplets, which could be explained by the fact that not all knowledge rules in fashion domain can be universally applicable to all the fashion item pairs.

 \begin{table}[!t]
\vspace{-1em}
\centering \caption{Effects of the rule guidance. The first row refers to the  performance of the baseline DBPR. }\label{table8}
\vspace{-1em}
  \begin{tabular}{|L{0.4cm}||*{1}{C{1.5cm}|C{2.4cm}|C{1.15cm}|C{1.15cm}|}}
     \hline
     Id & Top & Bottom &AUC-p&AUC-q\\\hline \hline
     0 & - & - &0.7704& -  \\\hline \Xhline{1.2pt}
     1& stripe & stripe &  0.7738 & 0.7738  \\\hline
     2& floral & floral &  0.7744 & 0.7739  \\\hline
     3& white & black & 0.7714 & 0.7714 \\\hline
     4& black & black & 0.7755 & 0.7770 \\\hline
     5& cashmere & leather &  0.7770 & 0.7773  \\\hline
     6& Yoins & Yoins & 0.7792 & 0.7790 \\\hline
     7& tank tops & shorts & 0.7732 & 0.7725 \\    \hline
     8& sweatshirt & activewear pants & 0.7757 & 0.7777 \\\hline
     9& coat & dress & 0.7794& 0.7792 \\\hline \Xhline{1.2pt}
     10& no silk & knit & 0.7739 & 0.7739 \\\hline
     11& no silk & chiffon & 0.7744 & 0.7744 \\\hline
     12& no coat & shorts & 0.7760 & 0.7755 \\\hline
     13 & no jacket & shorts& 0.7779 & 0.7779 \\\hline
     14& no blouses & dress & 0.7814 & 0.7814 \\\hline
     15& no T-shirt & dress & 0.7810 & 0.7815 \\\hline
   \end{tabular}
\end{table}

Moreover, to get a deep understanding of the rule guidance, we further conducted experiments on each rule. Table~\ref{table8} exhibits the performance of the student network and teacher network with different rules. Notably, we found that the negative rules (e.g., ``no T-shirt + dress'') seem to achieve better performance as compared to the positive ones (e.g., ``coat + dress''). One possible explanation is that people are more likely to distinguish the incompatible pairs than the compatible ones. In addition, as we can see, rules regarding category show superiority over rules pertaining to other attributes, such as material and color. This may be due to two reasons: 1) The category related rules are more specific and acceptable by the public, and hence have strong rule confidences and provide better guidance to the neural networks. 2) The category metadata is better structured, cleaner and more complete as compared to the loose and noisy title description, where we derived the other attributes (e.g., material and color) for fashion items. Moreover, as to the color related rules, we found that the rule ``black + black'' surprisingly outperforms the rule ``white + black''. One plausible explanation is that white tops are more versatile than black ones, suit more bottoms with different colors, and hence deteriorate the confidence of the rule ``white + black''. Last but not least, interestingly, we noted that the rule pertaining to the brand (i.e., Yoins) of fashion items can achieve remarkable performance. This may be due to that items of the same brand can share the brand exclusive features and hence are more likely to make suitable outfits.

\subsection{On Attention Mechanism (RQ2)}
To evaluate the importance of the attention mechanism in the knowledge distillation, we further compared AKD-DBPR with its derivation UKD-DBPR, where the rule confidence is assigned uniformly. Moreover, to obtain a thorough understanding, we conducted the comparative experiments with different modality configurations. Table~\ref{table7} shows the effects of the attention mechanism in our model with different modality combinations. First, as can be seen, our model consistently shows superiority over UKD-DBPR across different modality configurations, which enables us to  safely draw the conclusion that it is advisable to assign rule confidence attentively rather than uniformly. Second, we observed that AKD-DBPR remarkably outperforms DBPR with only the visual modality (the relative improvement reaches 6.97\%). This may be due to the fact that in this context, AKD-DBPR is able to take advantage of the contextual information to determine whether a sample satisfy the given rule, perform the knowledge distillation and hence significantly boost the performance of that solely with visual information. Moreover, we found that even with only the contextual modality, AKD-DBPR can achieve better performance than DBPR (the relative improvement is 2.69\%). One possible explanation is that the pure data-driven neural networks cannot accurately capture all the underlying matching rules with the limited labeled samples and thus need the domain knowledge to overcome this limitation.

\begin{table}[!t]
\vspace{-1em}
\centering \caption{Effects of the attention mechanism. }\label{table7}
\vspace{-1em}
\begin{tabular}{|L{2.6cm}||*{1}{C{1.2cm}|C{1.2cm}|C{1.2cm}|}}
     \hline
     Approaches & Text & Visual & All\\\hline \hline
   \textbf{AKD-DBPR-q}&$\mathbf{0.7374}$ &$\mathbf{0.7302}$ &$\mathbf{0.7852}$ \\
   \textbf{AKD-DBPR-p}&$0.7345$ &$0.6961$ &$0.7843$ \\\hline
   \textbf{UKD-DBPR-q}&$0.7245$ &$0.7280$ &$0.7760$ \\
   \textbf{UKD-DBPR-p}& $0.7275$ &$0.6865$ &$0.7785$ \\\hline
   \textbf{DBPR}& $0.7181$ &$0.6826$ &$0.7704$ \\\hline
   \end{tabular}
\end{table}

\begin{figure}[!b]
  \centering
  \includegraphics[scale=0.37]{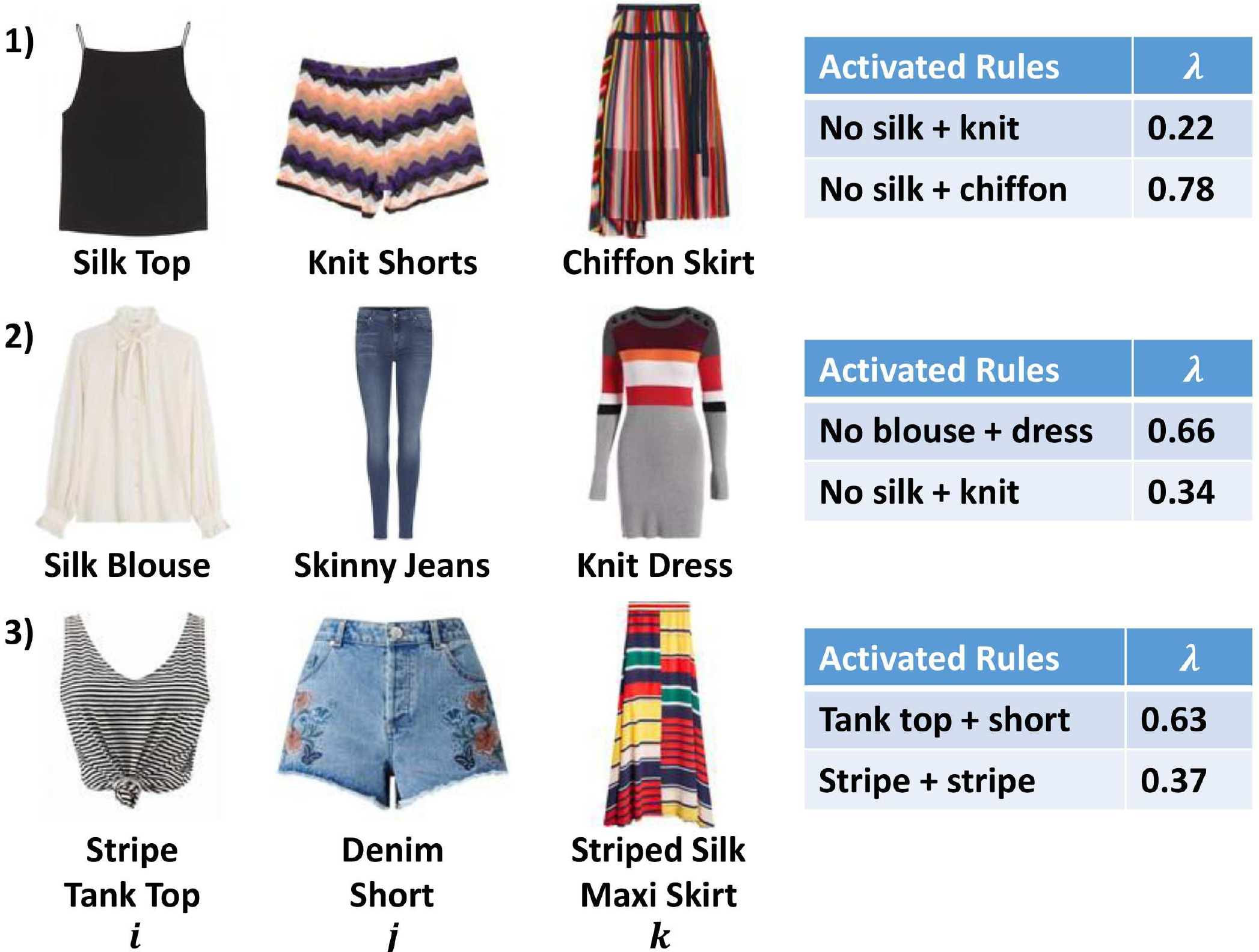}
  \vspace{-0.8em}
  \caption{Illustration of attentive rule confidences. }\label{fig6}
    \vspace{-1.2em}
\end{figure}

Apart from the quantitative analysis, we also provided certain intuitive examples to illustrate the effects of the attention mechanism in our scheme. Figure~\ref{fig6} illustrates  several examples regarding the rule confidence learned by the attention mechanism. As can be seen, different levels of rule confidence can be assigned for the same rule (``no silk + knit'') with different triplets. In addition, we found that the rules pertaining to the category are usually assigned higher confidence levels. This may be attributed to that people tend to put the category attribute at the first place when they make outfits compared to other attributes. Furthermore, we also noted that although the contextual metadata indicates that the third triplet activates the rule ``stripe + stripe'', the learned rule confidence is not much high. This may be due to the fact that the given rule is a bit fuzzy and general and the visual signals imply the incompatibility between the stripes in the given top $i$ and bottom $k$. Accordingly, to certain extent, the attention mechanism can be helpful to overcome the limitation of the human-defined fuzzy rules.

\begin{figure*}[!t]
  \centering
     \vspace{-1.5em}
   \subfigure[Observed testing tops]{
  \includegraphics[scale=0.46]{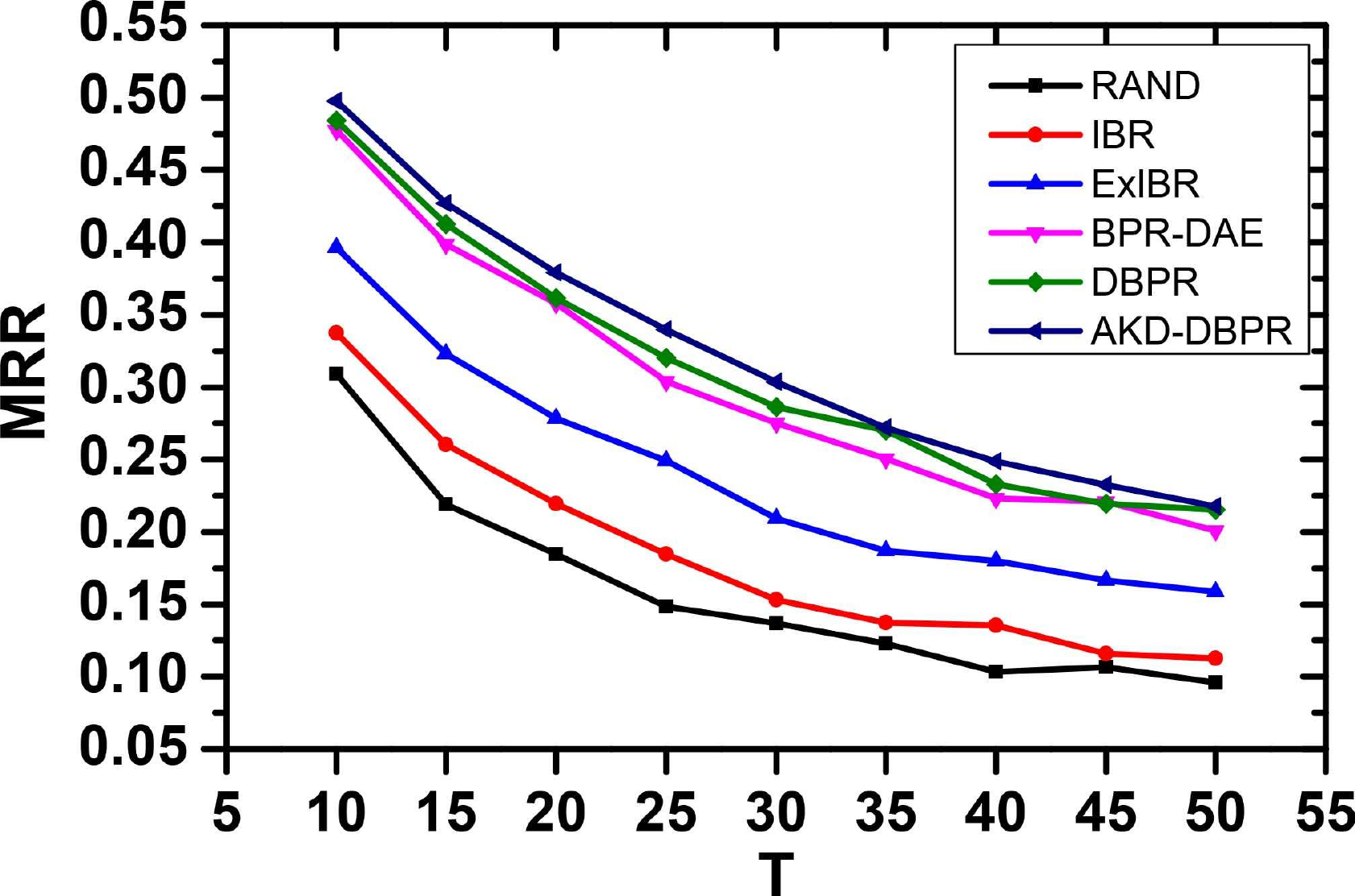}
   \label{fig8:subfig1}
   }
   \subfigure[Unobserved testing tops]{
  \includegraphics[scale=0.46]{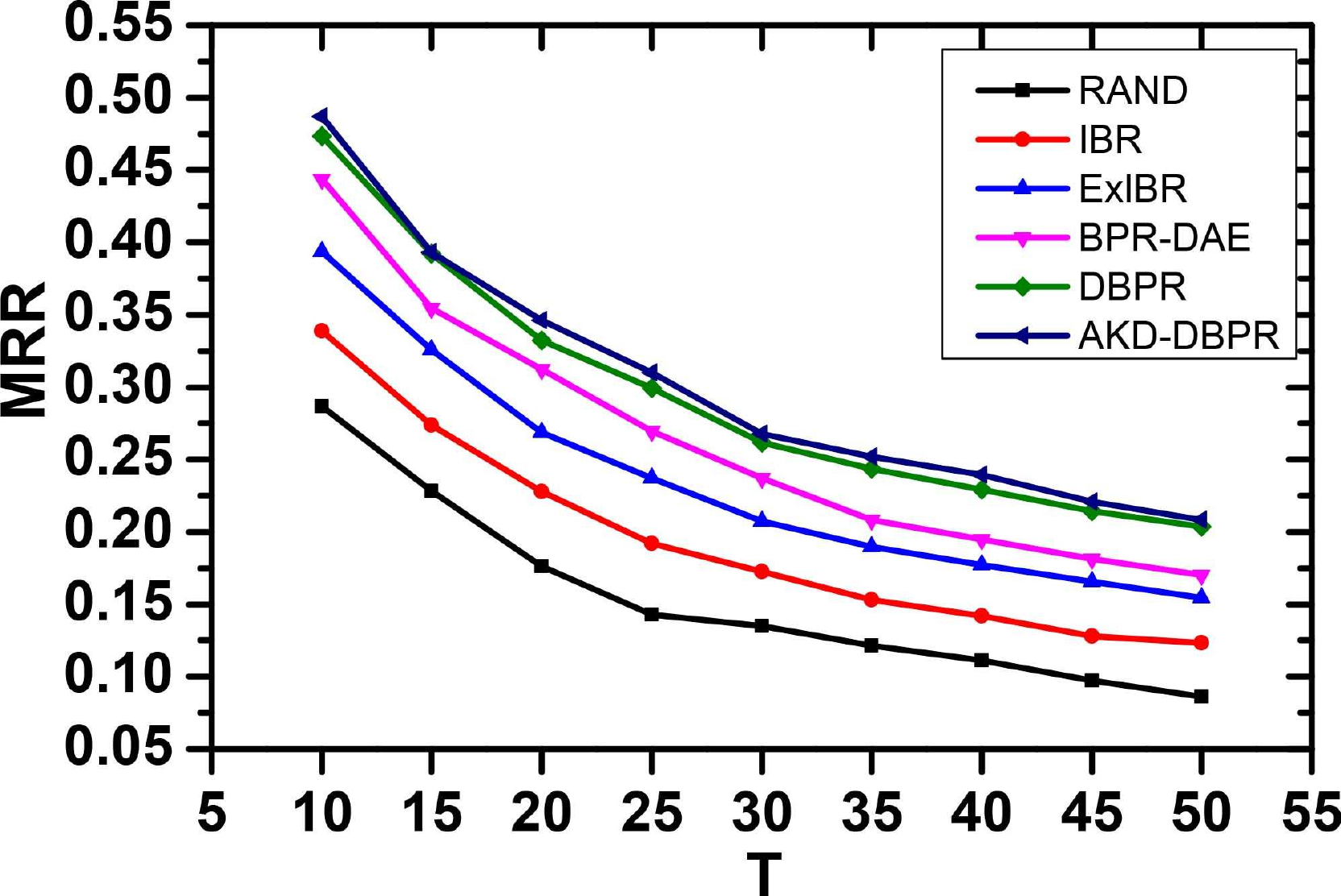}
   \label{fig8:subfig2}
   }
    \vspace{-1.2em}
  \caption{Performance of different models. }\label{fig7}
\end{figure*}


\begin{figure*}[!b]
  \centering
  \includegraphics[scale=0.66]{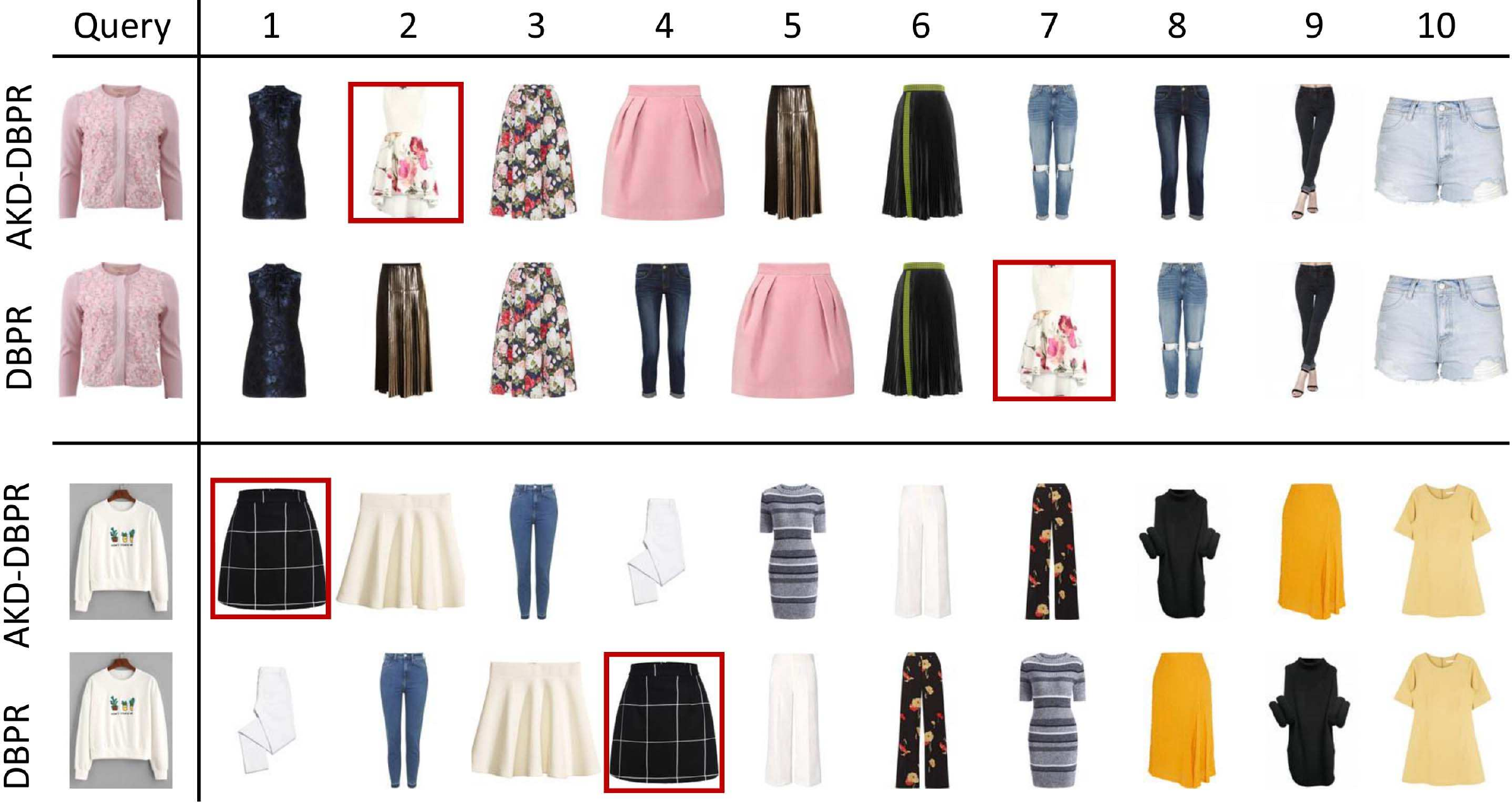}
  \vspace{-0.3em}
  \caption{Illustration of the ranking results of AKD-DBPR and DBPR. The bottoms highlighted in the red boxes are positive ones. The first example activates the rules ``floral + floral'' and ``coat + dress'', while the second example triggers the rule ``white + black''.}\label{fig8}
    \vspace{-1.6em}
\end{figure*}

\subsection{On Fashion Item Retrieval (RQ3)}
To assess the practical value of our work, we evaluate the proposed AKD-DBPR towards the complementary fashion item retrieval. As it is time-consuming to rank all the bottoms for each top, we adopted the common strategy~\cite{HeLZNHC17} that feeds each top $t_i$ appeared in $\mathcal{S}_{test}$ as a query, and randomly selected $T$ bottoms as the ranking candidates, where there is only one positive bottom. Thereafter, by passing them to the trained neural networks, getting their latent representations and calculating the compatibility score $m_{ij}$ according to  Eqn.($\ref{eq2}$), we generated a ranking list of these bottoms for the given top. In our setting, we focused on the average position of the positive bottom in the ranking list and thus adopted the  mean reciprocal rank (MRR) metric~\cite{jiang2015fast}.

In total, we have $1,954$ unique tops in the testing set. Due to the sparsity of the real-world dataset, we found there are $1,262$ (64.59\%) tops never appeared in $\mathcal{S}_{train}$.  To comprehensively evaluate the proposed model, we compared it with different models using different type of testing tops: observed testing tops and unobserved ones. As can be seen from Figure~\ref{fig7}, AKD-DBPR and DBPR outperform all the other baselines consistently at different numbers of bottom candidates in all scenarios, which demonstrates the effectiveness of our models in complementary fashion item retrieval. In addition, AKD-DBPR and DBPR achieve satisfactory performance with both observed and unobserved tops, which validates their capability of handling the cold start problem. Last but not least, we found that AKD-DBPR outperforms DBPR in both scenarios, especially with observed testing tops, which reconfirms the importance of incorporating the domain knowledge. To have an intuitive understanding of the results, we provided certain intuitive ranking results of AKD-DBPR and DBPR for testing tops in Figure~\ref{fig8}. The bottoms highlighted in the red boxes are the positive ones. By checking the context of each example, we found that they both activate certain matching rules, such as ``floral + floral'', ``coat + dress'' and ``white + black'', which may contribute to the good performance of AKD-DBPR.

\section{Conclusion and Future Work}
In this work, we present an attentive knowledge distillation scheme towards compatibility modeling in the context of clothing matching, which jointly learns from both the specific data samples and general knowledge rules. Considering that different rules can have different confidence levels to different samples, we seamlessly sew up the attention mechanism into the knowledge distillation framework to attentively assign the rule confidence. Extensive experiments have been conducted on the real-world dataset and the encouraging empirical results demonstrate the effectiveness of the proposed scheme and indicate the benifits of taking the domain knowledge into consideration in the context of compatibility modeling.  We find that the negative matching rules and category related rules seem to be more powerful than others. We also exhibited the benefits  of incorporating the attention mechanism into the knowledge distillation framework.

One limitation of our work is that currently we only rely on the contextual metadata to identify the rules activated by the given sample, which is largely constrained by the incomplete and noisy description. In the future, we plan to explore the potential of visual signals in the rule identification.
\begin{acks}
This work is supported by the National Basic Research Program of China (973 Program), No.: 2015CB352502; National Natural Science Foundation of China, No.: 61772310 and 61702300; and the Project of Thousand Youth Talents 2016.

\end{acks}

\bibliographystyle{ACM-Reference-Format}
\bibliography{sigproc_abbre}

\end{document}